\documentclass{svproc}

\usepackage{url}

\usepackage[detect-none,load-configurations=binary]{siunitx}
\usepackage[dvipsnames, table]{xcolor}
\usepackage{graphicx}
\usepackage{siunitx}
\DeclareSIUnit{\nothing}{\relax}
\usepackage{booktabs}
\usepackage[hidelinks]{hyperref}
\usepackage{multirow}
\usepackage{xcolor}
\usepackage{comment}

\begin{document}
\mainmatter              
\title{Fusing Multi-sensor Input with State Information on TinyML Brains for Autonomous Nano-drones}

\titlerunning{Fusing Multi-sensor on TinyML}  
\author{Luca Crupi\inst{1} \and Elia Cereda\inst{1} \and Daniele Palossi\inst{12}}
\authorrunning{Luca Crupi et al.}
\tocauthor{Luca Crupi, Elia Cereda, and Daniele Palossi}
\institute{Dalle Molle Institute for Artificial Intelligence, USI-SUPSI, Lugano, Switzerland,\\
\and
Integrate Systems Laboratory, ETH Z\"urich, Z\"urich, Switzerland\\
\email{name.surname@idsia.ch}}

\maketitle             

\begin{abstract}
Autonomous nano-drones ($\sim$\SI{10}{\centi\meter} in diameter), thanks to their ultra-low power TinyML-based brains, are capable of coping with real-world environments.
However, due to their simplified sensors and compute units, they are still far from the sense-and-act capabilities shown in their bigger counterparts.
This system paper presents a novel deep learning-based pipeline that fuses multi-sensorial input (i.e., low-resolution images and 8$\times$8 depth map) with the robot's state information to tackle a human pose estimation task.
Thanks to our design, the proposed system -- trained in simulation and tested on a real-world dataset -- improves a state-unaware State-of-the-Art baseline by increasing the $R^2$ regression metric up to 0.10 on the distance's prediction.

\end{abstract}

\section{Introduction} \label{sec:intro}

Miniaturized autonomous nano-drones, from sub-\SI{30}{\gram} blimps~\cite{palossi2017self} to palm-sized quadrotors~\cite{crupi2023sim}, are becoming increasingly popular in academia and industry thanks to their broad applicability~\cite{uav_survey_applications}. 
They can assist human operators in rescue missions, inspect narrow places, and monitor indoor facilities.
Additionally, thanks to their small footprint and weight, they can safely operate close to people, being harmless.
However, the vast versatility of autonomous nano-drones comes at the price of simple and cheap electronics, leading to ultra-constrained computational resources and simplistic sensors, e.g., low-resolution cameras.

Recent works~\cite{crupi2023sim,kaufmann2019beauty,zhang2024end} have shown how, despite their limitations, nano-drones can achieve a high level of intelligence using tiny deep learning (DL) models fed with multiple sensors and relying only on sub-\SI{100}{\milli\watt} System-on-Chips (SoCs).
The vast majority of these works address DL-based perception tasks focusing on understanding the surrounding environment, such as estimating the relative pose of an object or a human~\cite{crupi2023sim,kaufmann2019beauty} or avoiding obstacles~\cite{LIU20239312,zhang2024end}.
We define these as \textit{allocentric} tasks, in which the prediction refers to an external object/subject, in contrast to \textit{egocentric} tasks, in which the subject is the robot itself.

In DL models for egocentric tasks, such as state estimation and visual odometry~\cite{vinet,deepvio,7927257}, exploiting the robot's state as an auxiliary input is common.
However, whether it can also benefit allocentric tasks is still an open research question, which we aim to explore by introducing state information into a convolutional neural network (CNN) for the allocentric task of human pose estimation.
We start from a State-of-the-Art (SoA) baseline CNN, running on a Greenwaves Technologies (GWT) GAP8 SoC and fed with 160$\times$96 pixel images and an 8$\times$8 depth map, both acquired aboard a nano-drone and used to estimate the distance ($x$) and the lateral displacement ($y$) between the drone and a human in front of it.
Then, we extend this model by feeding part of the nano-drone's state as an additional input of the CNN, and we present an ablation study exploring \textit{i}) how different state information (i.e., only pitch or roll, and pitch+roll) and \textit{ii}) how different fusion techniques affect the final CNN's regression performance.

Our exploration leads to four alternative fusion approaches, resulting in just as many CNNs we deploy and characterize on the GAP8 SoC aboard our target nano-drone.
Despite the CNNs training being done only in simulation, our results from a real-world test set (more than \SI{3.5}{\kilo\nothing} image-depth pairs) highlight how the introduction of either of the state angles improves the regression performance of our neural network, measured with an $R^2$ increase of +0.10 on $x$, and +0.02 on $y$, at an increase of the computational cost, measured with the number of multiply-and-accumulate (MAC) operations, of just 0.11\% if compared with the state-unaware SoA baseline model~\cite{crupi2023sim}.
\section{System description} \label{sec:system}

\begin{figure}[t]
  \centering
  \includegraphics[width=\textwidth]{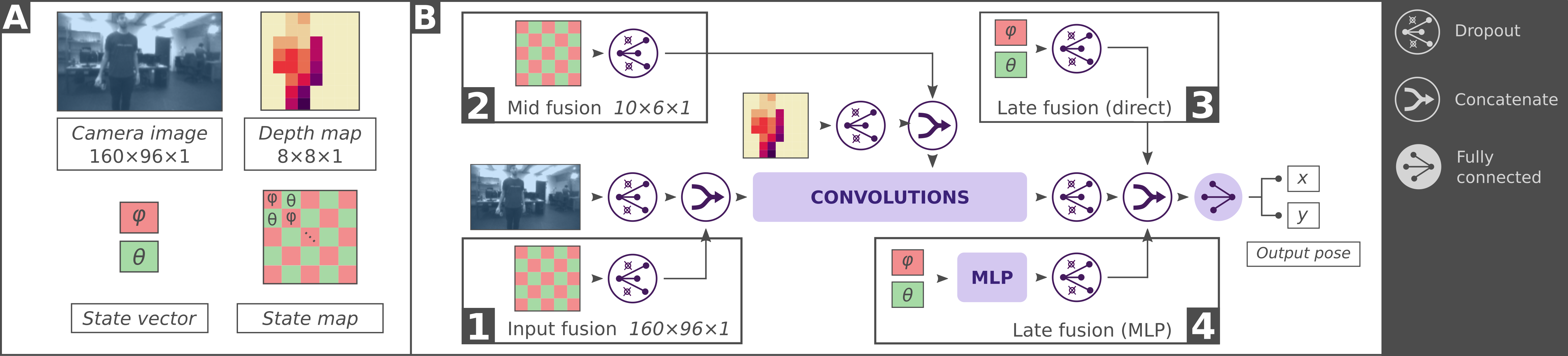}
  \caption{A) Our CNN inputs. B) CNN architecture exploration, based on the SoA vision+depth backbone from~\cite{crupi2023sim}. Our proposed state-aware models either with 1) input fusion, 2) mid fusion, 3) late fusion (direct), and 4) late fusion with MLP.}
  \label{fig:model}
\end{figure}

Our work starts from the state-unaware vision+depth CNN presented in~\cite{crupi2023sim} for the human pose estimation task.
We share the same target nano-drone, i.e., the Bitcraze Crazyflie 2.1, and the same computational and sensorial resources, such as the STM32 microcontroller, the GWT GAP8 octa-core SoC, the Himax HM01B0 monocular grayscale QVGA camera, and the ST VL53LC5CX Time-of-Flight (ToF) multi-zone ranging sensor.
Like in~\cite{crupi2023sim}, we process grayscale $160\times96$ images and 8$\times$\SI{8}{px} depth maps, while we predict the $(x, y)$ position coordinates of the human subject, as shown in Fig.~\ref{fig:model}.
The depth information is fused with a \textit{mid fusion} approach (i.e., the best-performing in~\cite{crupi2023sim}) that introduces the depth map as one of the feature maps in input to the CNN's 6$^{th}$ convolutional layer.

Our state-aware CNNs use the \textit{roll} ($\varphi$) and \textit{pitch} ($\theta$) components of the drone attitude, with the two alternative representations shown in Fig.~\ref{fig:model}-A: either a two-dimensional \textit{state map} in which the elements repeat in a Bayer-like pattern, or a two-element ($\varphi, \theta$) \textit{state vector}.
The former representation is more suitable for fusion approaches in which the state should be spread across multiple neural connections, such as in Fig.~\ref{fig:model}-B1 called \textit{input fusion}, and in Fig.~\ref{fig:model}-B2 called \textit{mid fusion}.
The sizes of the two-state maps vary to match each convolutional layer's input dimensions, i.e., $160\times96$ and $10\times6$.
Employing a two-element state vector, instead, is a cheaper solution -- i.e., less memory and MAC operations -- suitable for introducing the state directly into the last CNN's fully connected layer, depicted in Fig.~\ref{fig:model}-B3 and called \textit{late fusion (direct)}, or after processing it with a 2-layer multilayer perceptron (MLP), Fig.~\ref{fig:model}-B4.
As in~\cite{crupi2023sim}, for all variants, input dropout is employed at training time to use all three inputs equally, i.e., camera, depth, and state.

Finally, we train and validate the four models with the same \SI{700}{\kilo\nothing}-sample dataset (from the Webots simulator) and procedures as in~\cite{crupi2023sim}. 
We apply domain randomization, varying the environment (walls, floors, and decoy objects) and the human subject (chosen among 27 3D models).
Our procedure randomly selects the drone's attitude for each sample, such that our state variables (roll and pitch) are uniformly distributed over the $[\ang{-20}, \ang{+20}]$ range.
All CNNs are trained with stochastic gradient descent at a learning rate of 0.001 for 100 epochs, selecting the model with the lowest validation loss.
We evaluate the models' performance on a challenging 3500-sample real-world test set, collected in a room equipped with a motion capture system, where the drone flights stress roll and pitch in the same $[\ang{-20}, \ang{+20}]$ range. 

\section{Experimental results} \label{sec:results}

\begin{figure}[t]
    \centering
    \resizebox{\columnwidth}{!}{%
    \includegraphics{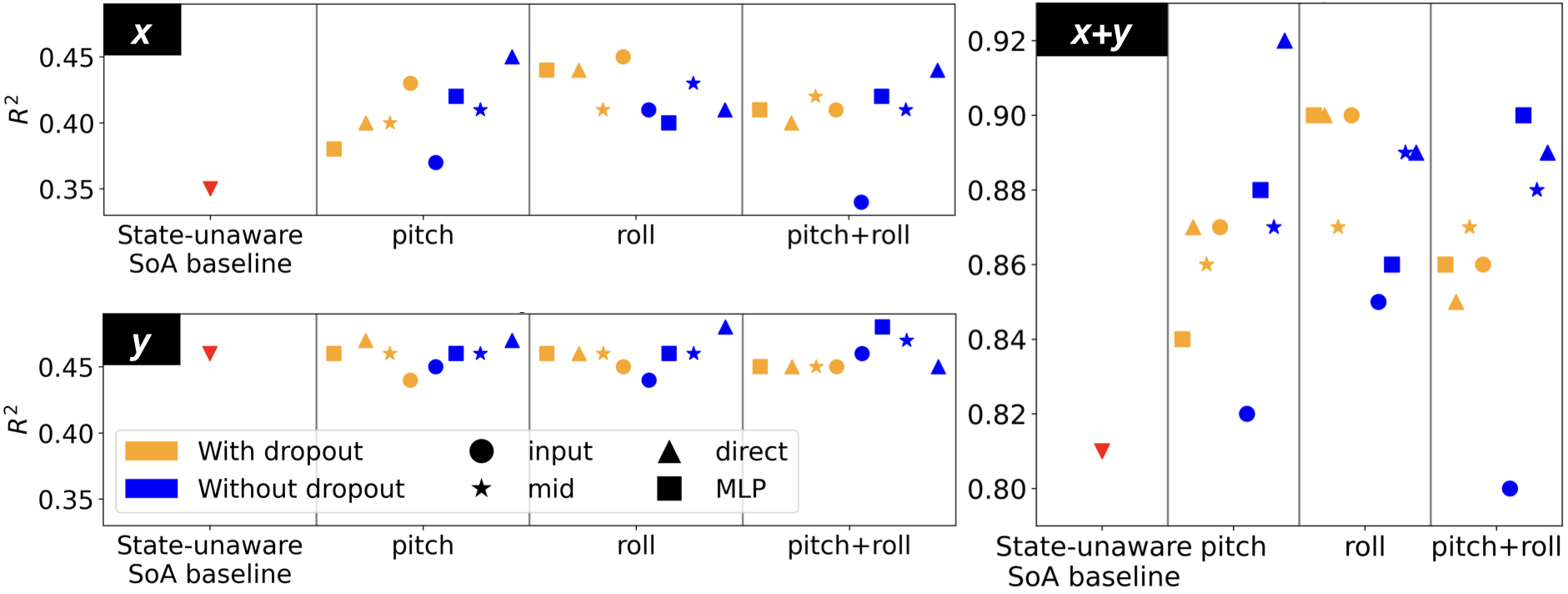}}
    \caption{$R^2$ comparison of our fusion techniques (i.e., input, mid, direct, and MLP) with and without dropout and with different states (i.e., pitch, roll, and pitch+roll) vs. the state-unaware SoA baseline~\cite{crupi2023sim}. Each marker is the average of 5 different training.}
    \label{fig:resultsR2}
\end{figure}

In this section, we present our ablation study on the proposed fusion techniques, i.e., input, mid, and two late fusions (MLP and direct), combining them with different states (i.e., pitch, roll, pitch+roll) and the effect of a dropout layer applied on the state during training.
Each combination results in a different model (24 in total) compared against the state-unaware SoA baseline proposed in~\cite{crupi2023sim}.
Each model is trained 5 times, resulting in 125 distinctive instances, presented in Fig.~\ref{fig:resultsR2} and evaluated with the $R^2$ score on the $x$ and $y$ output variables.
The $R^2$ score measures regression performance in a normalized range [$-\infty$, 1]. 
A perfect predictor scores 1, while a dummy predictor that always predicts the average of the test set would score 0.

The results in Fig.~\ref{fig:resultsR2} highlight that introducing any state is consistently beneficial (with the only exception of input fusion without dropout and pitch+roll states), regardless of dropout and fusion technique, with the sum of the $R^2$ improvements on $x$ and $y$ up to +0.11 w.r.t. the state-unaware SoA baseline model.
Applying the dropout on the state input does not affect the $R^2$ standard deviation, contrary to the results from~\cite{crupi2023sim}.
In this regard, input dropout in~\cite{crupi2023sim} was used on the vision and depth sensorial inputs to force the network to consider both information streams.
Instead, in this case, the model does not overfit on the state even without dropout because it would be insufficient to estimate the person's position without a camera or ToF sensor. 
Out of all CNNs, the best-performing model uses only the pitch state, without dropout and with late fusion (direct). 
Its $R^2$ increases by +0.10 on the $x$ and by +0.01 on the $y$ output w.r.t. the state-unaware SoA baseline, with a negligible increase in the computational cost, i.e., 4 MAC.
The configuration input fusion on pitch+roll without dropout is the only one where the performance slightly decreases on $x$ (-0.01 $R^2$).

\begin{table}[b!]
    \tiny
    \centering
    \resizebox{\columnwidth}{!}{%
    \begin{tabular}{l|ccccc}
    \toprule
    \textbf{}&SoA baseline & Input & Mid & Direct & MLP\\
    \midrule
    \textbf{$R^2$ (peak $x$/$y$)} & 0.35/0.46 & 0.45/0.45 & 0.43/0.47 & 0.45/0.48 & 0.44/0.48\\
    \textbf{Memory [\SI{}{\kilo\byte}]} & 304.4 & 305.2 & 305.5 & 304.4 & 304.8 \\
    \textbf{MAC/frame [\SI{}{\mega\nothing}]} & 14.7 & 17.8 & 14.8 & 14.7 & 14.7\\
    \bottomrule
    \end{tabular}
    }
    \caption{$R^2$ peak performance, memory occupancy, and computational cost for the state-unaware SoA baseline and the four different state fusion techniques. We report the maximum $R2$ for each variable, $x$ and $y$, and for each fusion method.}
    \label{tab:mem_mmac}
\end{table} 

Tab.~\ref{tab:mem_mmac} further reports the $R^2$ peak performance for each output ($x$ and $y$ from different models), the memory occupancy, and the computational cost for each state fusion technique and for the state-unaware SoA baseline.
Among the four architecture variants, input fusion and direct late fusion achieve the highest regression performance improvements (+0.10 and +0.11, respectively) w.r.t the SoA baseline.
However, input fusion is the most expensive in computational and second most expensive in memory cost, respectively +\SI{3.1}{\mega MAC/frame} (+21\%) and  +\SI{0.8}{\kilo\byte} (+2.6\%) w.r.t. the baseline.
The other three variants (i.e., mid, MLP, and direct fusion) increase the MAC workload by less than 0.11\%. 
However, the two late-fusion variants require less memory compared to mid-fusion (+1.3\% increase vs. +3.6\%, respectively).
Thus, considering the trade-off between regression performance and memory and computational costs suggests that the direct late-fusion model is the most convenient.
These results highlight the benefit provided by the fusion of the state in the pose estimation task with minimal or no computational cost addition (between 0 and 0.11\% more MAC for the mid, direct, and MLP approaches) and with a limited increase of the memory usage, i.e., between 0 and 0.36\% w.r.t. the SoA baseline.

\section{Conclusion} \label{sec:conclusion}

This work introduces a novel study on how to fuse state information into a lightweight multi-sensorial (camera and depth map) CNN for the allocentric task of human pose estimation aboard an autonomous nano-drone.
Our analysis highlights the impact on the regression performance of different states (roll, pitch, roll+pitch) and fusion techniques (input, mid-fusion, and late fusions).
Our key findings consistently show the benefit of the state fusion: mean $R^2$ improvement of 0.06 on the $x$ variable w.r.t. the SoA baseline model.
Our best model, the late fusion approach, increases the $R^2$ up to 0.10 and 0.01 on $x$ and $y$ with a negligible overhead in memory or computation.

\bibliographystyle{abbrv}
\bibliography{bibliography}

\end{document}